\documentclass[11pt]{article}
\usepackage{coling2016}
\usepackage{times}
\usepackage{url}

\usepackage{booktabs}
\usepackage{multirow}
\usepackage{caption}
\usepackage{subcaption}
\usepackage{graphicx}
\usepackage{amsthm}
\usepackage{amsfonts}
\usepackage{array}
\usepackage{enumitem}
\usepackage[font={small}]{caption}
\theoremstyle{definition}

\title{Semi-supervised Word Sense Disambiguation with Neural Models}
\begin{document}
\author{Dayu Yuan \qquad
  Julian Richardson \qquad
  Ryan Doherty \qquad
  Colin Evans \qquad
  Eric Altendorf \\
   Google, Mountain View CA, USA \\
   \{dayuyuan,jdcr,portalfire,colinhevans,ealtendorf\}@google.com}
\maketitle
\begin{abstract}
Determining the intended sense of words in text -- word sense disambiguation (WSD) -- is a long-standing problem in natural language processing. Recently, researchers have shown promising results using word vectors extracted from a neural network language model as features in WSD algorithms. However, a simple average or concatenation of word vectors for each word in a text loses the sequential and syntactic information of the text. 
In this paper, we study WSD with a sequence learning neural net, LSTM, to better capture the sequential and syntactic patterns of the text. To alleviate the lack of training data in all-words WSD, we employ the same LSTM in a semi-supervised label propagation classifier. We demonstrate state-of-the-art results, especially on verbs. 
\end{abstract}

\section{Introduction}
\blfootnote
{
    \hspace{-0.65cm}
    This work is licensed under a Creative Commons 
    Attribution 4.0 International License.
    License details: \url{http://creativecommons.org/licenses/by/4.0/}
}
Word sense disambiguation (WSD) is a long-standing problem in natural language processing (NLP) with broad applications. Supervised, unsupervised, and knowledge-based approaches have been studied for WSD~\cite{survey}. However, for {\it all-words} WSD, where all words in a corpus need to be annotated with word senses, it has proven extremely challenging to beat the strong baseline, which always assigns the most frequent sense of a word without considering the context~\cite{semeval2007,survey,semeval2013,semeval2015}. Given the good performance of published supervised WSD systems when provided with significant training data on specific words~\cite{IMS}, it appears lack of sufficient labeled training data for large vocabularies is the central problem.

One way to leverage unlabeled data is to train a neural network language model (NNLM) on the data. Word embeddings extracted from such a NNLM (often Word2Vec~\cite{mikolov2013efficient}) can be incorporated as features into a WSD algorithm. Iacobacci et al.~\shortcite{supervised_word2vec} show that this can substantially improve WSD performance and indeed that competitive performance can be attained using word embeddings alone. 

In this paper, we describe two novel WSD algorithms. The first is based on a Long Short Term Memory (LSTM)~\cite{hochreiter1997long}. Since this model is able to take into account word order when classifying, it performs significantly better than an algorithm based on a continuous bag of words model (Word2vec)~\cite{mikolov2013efficient,supervised_word2vec}, especially on verbs. 

We then present a semi-supervised algorithm which uses label propagation~\cite{talukdar2009new,ravi2016} to label unlabeled sentences based on their similarity to labeled ones. This allows us to better estimate the distribution of word senses, obtaining more accurate decision boundaries and higher classification accuracy. 

The best performance was achieved by using an LSTM language model with label propagation. Our algorithm achieves state-of-art performance on many SemEval all-words tasks. It also outperforms the most-frequent-sense and Word2Vec baselines by $10$\% (see Section~\ref{sec:noad-evel} for details).

{\it Organization: } We review related work in Section~\ref{sec:related-work}. We introduce our supervised WSD algorithm in Section~\ref{sec:supervised-wsd}, and the semi-supervised WSD algorithm in Section~\ref{sec:semi-supervised-wsd}. Experimental results are discussed in Section~\ref{sec:experiment}. We provide further discussion and future work in Section~\ref{sec:discussion}.

\section{Related Work}
\label{sec:related-work}
The development of large lexical resources, such as WordNet~\cite{wordnet} and BabelNet~\cite{babelnet}, has enabled knowledge-based algorithms which show promising results on all-words prediction tasks~\cite{knowledge-rich,semeval2013,semeval2015}. 
WSD algorithms based on supervised learning are generally believed to perform better than knowledge-based WSD algorithms, but they need large training sets to perform well~\cite{semeval2007,semeval2007-coarse,survey,IMS}. 
Acquiring large training sets is costly. 
In this paper, we show that a supervised WSD algorithm can perform well with  $\sim 20$ training examples per sense. 

In the past few years, much progress has been made on using neural networks to learn word embeddings~\cite{mikolov2013efficient,levy2014neural}, to construct language models~\cite{mikolov2011extensions}, perform sentiment analysis~\cite{socher2013recursive}, machine translation~\cite{sequence_to_sequence} and many other NLP applications. 

A number of different ways have been studied for using word embeddings in WSD. There are some common elements:
\begin{itemize}[noitemsep,nolistsep] 
\item Context embeddings. Given a window of text $w_{n-k}, ..., w_n, ..., w_{n+k}$ surrounding a focus word $w_n$ (whose label is either known in the case of example sentences or to be determined in the case of classification), an embedding for the context is computed as a concatenation or weighted sum of the embeddings of the words $w_i, i \ne n$. Context embeddings of various kinds are used in both~\cite{chen2014unified} and~\cite{supervised_word2vec}. 
\item Sense embeddings. Embeddings are computed for each word sense in the word sense inventory (e.g. WordNet). In~\cite{autoextend}, equations are derived relating embeddings for word senses with embeddings for undisambiguated words. The equations are solved to compute the sense embeddings. In~\cite{chen2014unified}, sense embeddings are computed first as weighted sums of the embeddings of words in the WordNet gloss for each sense. These are used in an initial bootstrapping WSD phase, and then refined by a neural network which is trained on this bootstrap data.
\item Embeddings as SVM features. Context embeddings~\cite{supervised_word2vec,wsd-word-embedding}, or features computed by combining context embeddings with sense embeddings~\cite{autoextend}, can be used as additional features in a supervised WSD system e.g. the SVM-based {\em IMS}~\cite{IMS}. Indeed Iacobacci et al.~\shortcite{supervised_word2vec} found that using embeddings as the only features in {\em IMS} gave competitive WSD performance. 
\item Nearest neighbor classifier. Another way to perform classification is to find the word sense whose sense embedding is closest, as measured by cosine similarity, to the embedding of the classification context. This is used, for example, in the bootstrapping phase of~\cite{chen2014unified}.
\item Retraining embeddings. A feedforward neural network can be used to jointly perform WSD and adjust embeddings ~\cite{chen2014unified,wsd-word-embedding}.
\end{itemize}  

In our work, we start with a baseline classifier which uses $1000$-dimensional embeddings trained on a $100$ billion word news corpus using Word2Vec~\cite{mikolov2013efficient}. The vocabulary consists of the most frequent $1,000,000$ words, without lemmatization or case normalization. Sense embeddings are computed by averaging the context embeddings of sentences which have been labeled with that sense. To classify a word in a context, we assign the word sense whose embedding has maximal cosine similarity with the embedding of the context. This classifier has similar performance to the best classifier in~\cite{supervised_word2vec} when SemCor is used as a source of labeled sentences. The Word2Vec embeddings are trained using a bag of words model, i.e. without considering word order in the training context, and word order is also not considered in the classification context. In Section~\ref{sec:supervised-wsd} we show that using a more expressive language model which takes account of word order yields significant improvements.

Semi-supervised learning has previously been applied successfully to word sense disambiguation. In \cite{yarowsky1995unsupervised} bootstrapping was used to learn a high precision WSD classifier. A low recall classifier was learned from a small set of labeled examples, and the labeled set then extended with those sentences from an unlabeled corpus which the classifier could label with high confidence. The classifier was then retrained, and this iterative training process continued to convergence. Additional heuristics helped to maintain the stability of the bootstrapping process. The method was evaluated on a small data set. 

In \cite{niu2005word}, a label propagation algorithm was proposed for word sense disambiguation and compared to bootstrapping and a SVM supervised classifier. Label propagation can achieve better performance because it assigns labels to optimize a {\em global} objective, whereas bootstrapping propagates labels based on {\em local} similarity of examples. 

In Section~\ref{sec:semi-supervised-wsd} we describe our use of label propagation to improve on nearest neighbor for classification.

\section{Supervised WSD with LSTM}

\begin{figure}[ht]
    \centering
    \includegraphics[width=0.6\textwidth]{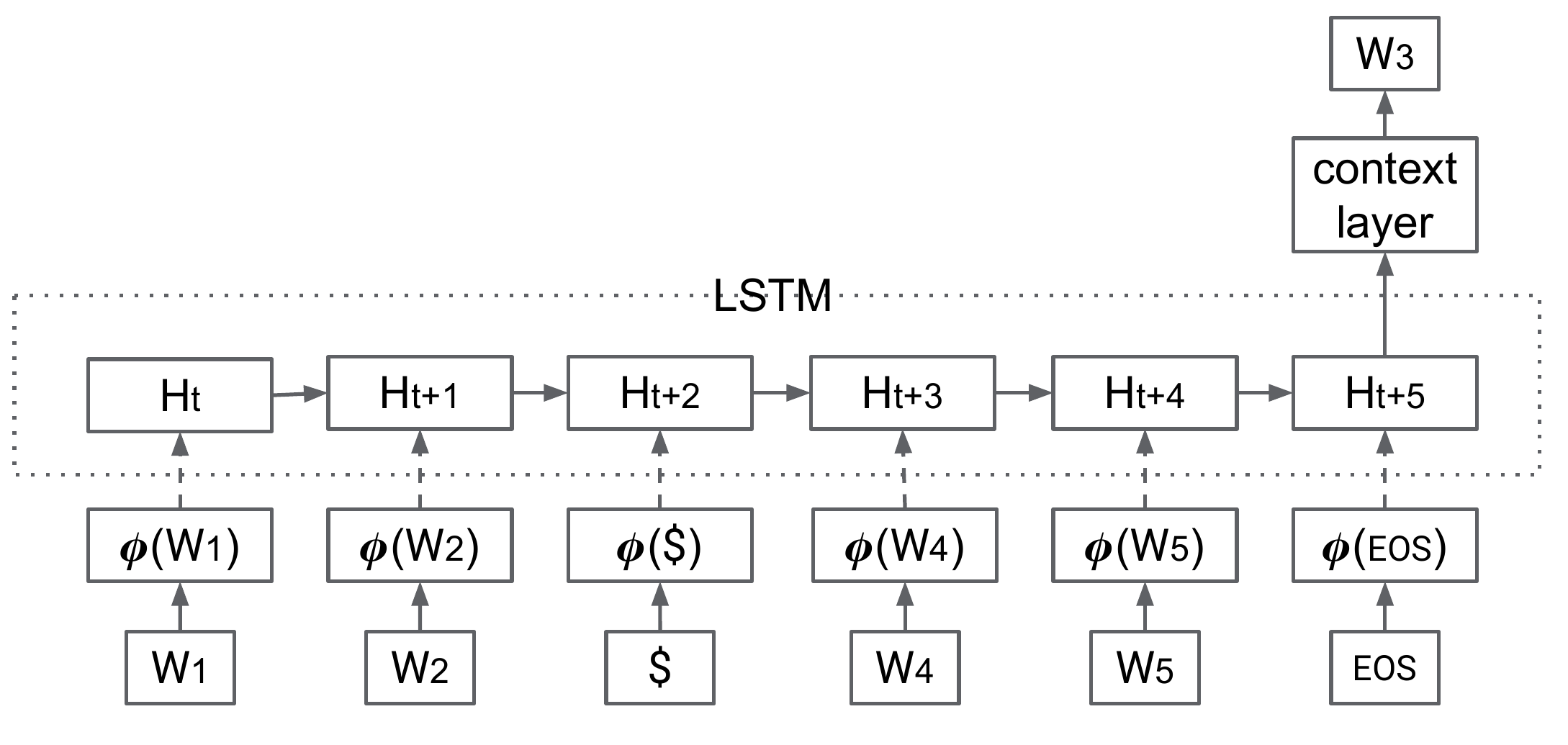}
    \caption{LSTM: Replace the focus word $w_3$ with a special symbol \$ and predict $w_3$ at the end of the sentence.}
    \label{fig:lstm}
\end{figure}

Neural networks with long short-term memory (LSTM) units~\cite{hochreiter1997long} make good language models which take into account word order~\cite{sundermeyer2012lstm}. We train a LSTM language model to predict the held-out word in a sentence. As shown in Figure~\ref{fig:lstm}, we first replace the held-out word with a special symbol \$, and then, after consuming the remaining words in the sentence, project the $h$ dimensional hidden layer to a $p$ dimensional context layer, and finally predict the held out word with softmax. By default, the LSTM model has $2048$ hidden units, $512$ dimensional context layer and $512$ dimensional word embeddings. We also studied other settings, see Section~\ref{sec:exp-noad-lstm} for details. We train the LSTM on a news corpus of about $100$ billion tokens, with a vocabulary of $1,000,000$ words. Words in the vocabulary are neither lemmatized nor case normalized. 

Our LSTM model is different from that of Kågebäck and Salomonsson~\cite{bilstm}. We train a LSTM language model, which predicts a held-out word given the surrounding context, with a large amount of unlabeled text as training data. The huge training dataset allows us to train a high-capacity model ($2048$ hidden units, $512$ dimensional embeddings), which achieves high precision without overfitting. In our experiments, this directional LSTM model was faster and easier to train than a bidirectional LSTM, especially given our huge training dataset. 
Kågebäck and Salomonsson's LSTM directly predicts the word senses and it is trained with a limited number of word sense-labeled examples. Although regularization and dropout are used to avoid overfitting the training data, the bidirectional LSTM is small with only $74 + 74$ neurons and $100$ dimensional word embeddings~\cite{bilstm}. 
Because our LSTM is generally applicable to any word, it achieves high performance on {\em all-words WSD} tasks (see Section~\ref{sec:experiment} for details), which is the focus of this paper. Kågebäck and Salomonsson's LSTM is only evaluated on {\em lexical sample WSD} tasks of SemEval 2 and 3~\cite{bilstm}.  

\label{sec:supervised-wsd}

The behavior of the LSTM can be intuited by its predictions. Table~\ref{tab:LSTM-predictions} shows the top $10$ words predicted by an LSTM language model for the word `stock' in sentences containing various senses of `stock'.

\begin{table}[h]
\centering
\small
\begin{tabular}{>{\centering\arraybackslash} m{0.5cm}m{7cm}m{5.5cm} >{\centering\arraybackslash} m{0.8 cm}}
\toprule
id & sentence & top 10 predictions from LSTM & sense  \\\hline
1 &  Employee compensation is offered in the form of cash and/or {\em stock}. &cash, stock, equity, shares, loans, bonus, benefits, awards, equivalents, deposits &\multirow{2}{0.8cm}{sense\#1} \\
\cmidrule(lr){1-3} 
2 & The {\em stock} would be redeemed in five years, subject to terms of the company's debt. & bonds, debt, notes, shares, stock, balance, securities, rest, Notes, debentures & {}\\\hline
3 &  These stores sell excess {\em stock} or factory overruns .  &inventory, goods, parts, sales, inventories, capacity, products, oil, items, fuel& sense\#2 \\\hline
4 &  Our soups are cooked with vegan {\em stock} and seasonal vegetables. & foods, food, vegetables, meats, recipes, cheese, meat, chicken, pasta, milk& sense\#3 \\\hline
query &  In addition, they will receive {\em stock} in the reorganized company, which will be named Ranger Industries Inc. & shares, positions, equity, jobs, awards, representation, stock, investments, roles, funds& ? \\\hline
\end{tabular}
\caption{Top predictions of `stock'  in 5 sentences of different word senses}\label{tab:LSTM-predictions}
\end{table}

\begin{figure}[ht]
    \centering
    \begin{subfigure}[b]{0.48\textwidth}
        \centering
        \includegraphics[width=0.5\textwidth]{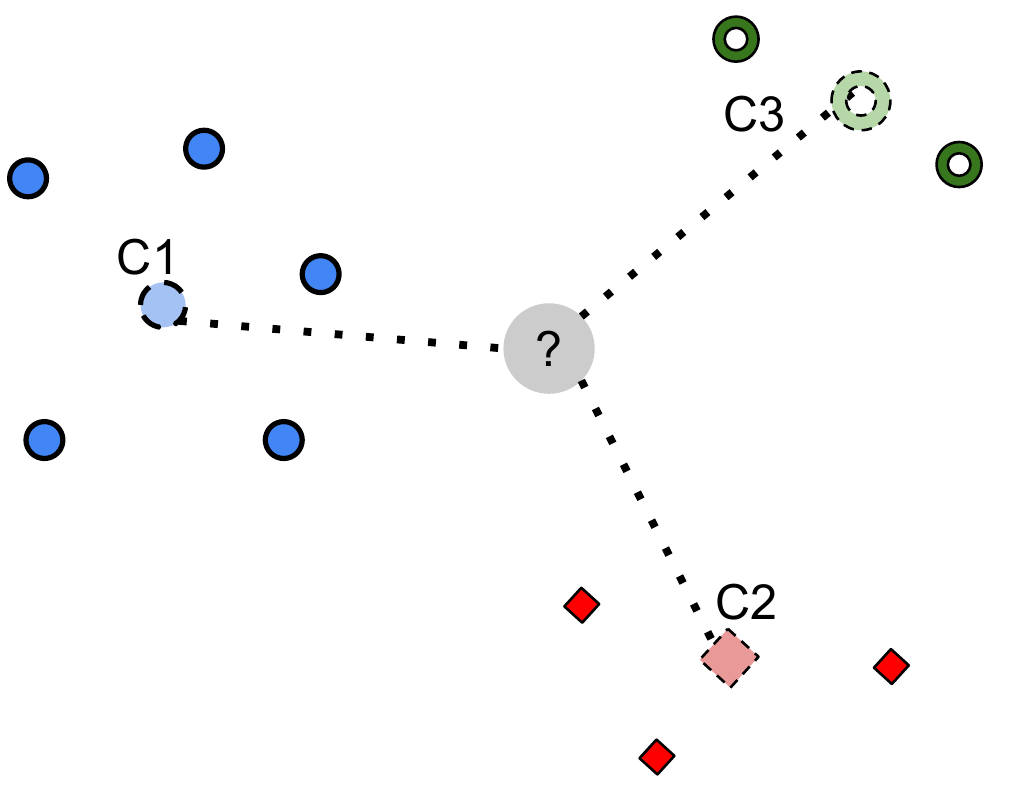}
        \caption{nearest neighbor}
        \label{fig:nneighbor_exp}
    \end{subfigure}
    \hfill
    \begin{subfigure}[b]{0.48\textwidth}
        \centering
        \includegraphics[width=0.5\textwidth]{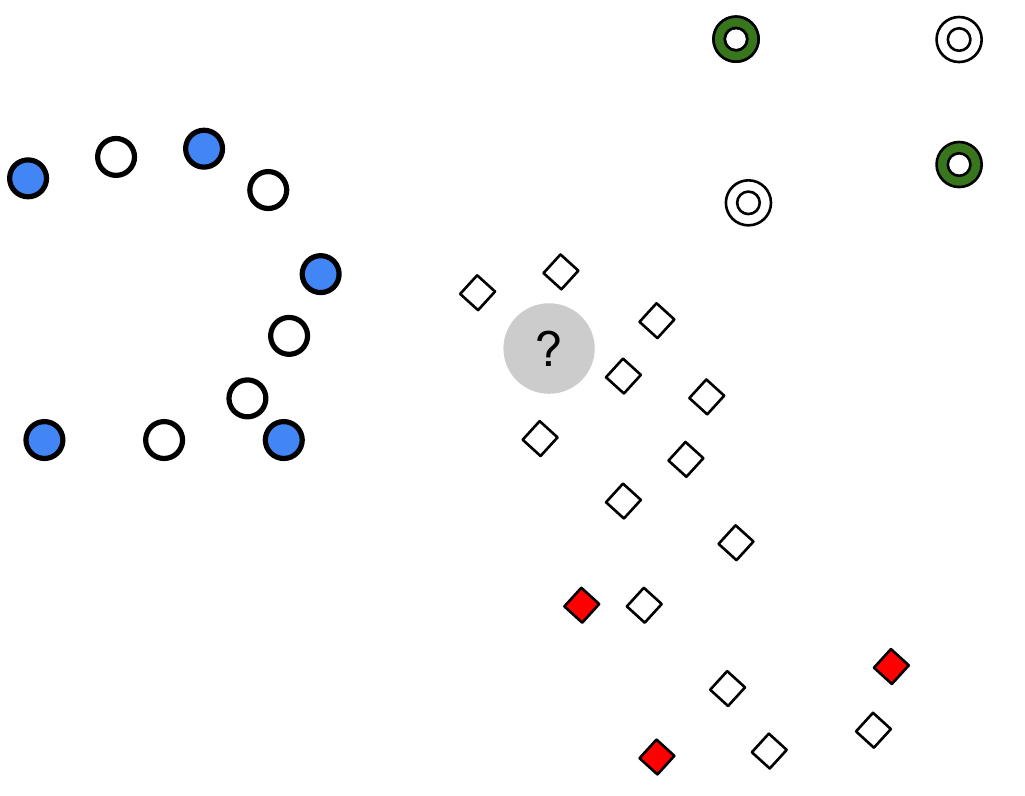}
        \caption{semi-supervised WSD with label propagation}
        \label{fig:expander_exp}
    \end{subfigure}
    \hfill
    \caption{WSD classifiers. Filled nodes represent labeled sentences. Unfilled nodes represent unlabeled sentences.}
    \label{fig:expander_exp}
\end{figure}

In our initial experiments, we computed similarity between two contexts by the overlap between their bags of predicted words. For example (Table~\ref{tab:LSTM-predictions}) the top predictions for the query overlap most with the LSTM predictions for `sense\#1' ---we predict that `sense\#1' is the correct sense. 
This bag of predictions, while easily interpretable, is just a discrete approximation to the internal state of the LSTM when predicting the held out word. We therefore directly use the LSTM's context layer from which the bag of predictions was computed as a representation of the context (see Figure~\ref{fig:lstm}).
Given context vectors extracted from the LSTM, our supervised WSD algorithms classify a word in a context by finding the sense vector which has maximum cosine similarity to the context vector (Figure~\ref{fig:nneighbor_exp}). We find the sense vectors by averaging context vectors of all training sentences of the same sense.
We observed in a few cases that the context vector is far from the held-out word's embedding, especially when the input sentence is not informative. For example, the LSTM language model will predict ``night" for the input sentence ``I fell asleep at [work]." when we hold out ``work". Currently, we treat the above cases as outliers. We would like explore alternative solutions, e.g., forcing the model to predict words that are close to one sense vector of the held-out word, in further work. 
As can be seen in SemEval all-words tasks and Tables~\ref{tab:baseline}, this LSTM model has significantly better performance than the Word2Vec models.

\section{Semi-supervised WSD}
\label{sec:semi-supervised-wsd}

The non-parametric nearest neighbor algorithm described in Section~\ref{sec:supervised-wsd} has the following drawbacks:
\begin{itemize}[noitemsep,nolistsep] 
\item
It assumes a spherical shape for each sense cluster, being unable to accurately model the decision boundaries given the limited number of examples. 
\item
It has no training data for, and does not model, the sense prior, omitting an extremely powerful potential signal. 
\end{itemize}

To overcome these drawbacks we present a semi-supervised method which augments the labeled example sentences with a large number of unlabeled sentences from the web.  Sense labels are then propagated from the labeled to the unlabeled sentences. Adding a large number of unlabeled sentences allows the decision boundary between different senses to be better approximated.

A {\em label-propagation graph} consists of (a) vertices with a number of labeled seed nodes and (b) undirected weighted edges. Label propagation (LP)~\cite{talukdar2009new} iteratively computes a distribution of labels on the graph's vertices to minimize a weighted combination of:
\begin{itemize}[noitemsep,nolistsep] 
\item The discrepancy between seed labels and their computed labels distributions.
\item The disagreement between the label distributions of connected vertices.
\item A regularization term which penalizes distributions which differ from the prior (by default, a uniform distribution).
\end{itemize}

We construct a graph for each lemma with labeled vertices for the labeled example sentences, and unlabeled vertices for sentences containing the lemma, drawn from some additional corpus.  Vertices for sufficiently similar sentences (based on criteria discussed below) are connected by an edge whose weight is the cosine similarity between the respective context vectors, using  the LSTM language model.
To classify an occurrence of the lemma, we create an additional vertex for the new sentence and run LP to propagate the sense labels from the seed vertices to the unlabeled vertices. 

Figure \ref{fig:expander_exp} (b) illustrates the graph configuration. Spatial proximity represents similarity of the sentences attached to each vertex and the shape of each node represents the word sense. Filled nodes represent seed nodes with known word senses. Unfilled nodes represent sentences with no word sense label, and the ? represents the word we want to classify. 

With too many edges, sense labels propagate too far, giving low precision. With too few, sense labels do not propagate sufficiently, giving low recall. We found that the graph has about the right density for common senses when we ranked vertex pairs by similarity and connected the pairs above the $95$ percentile. This may still leave rare senses sparsely connected, so we additionally added edges to ensure that every vertex is connected to at least $10$ other vertices. Our experiments (Table~\ref{tab:density}) show that this setting achieves good performance on WSD, and the performance is stable when the percentile ranges between $85$ to $98$.
Since it requires running LP for every classification, the algorithm is slow compared to the nearest neighbor algorithm. 

\section{Experiments}
\label{sec:experiment}
We evaluated the LSTM algorithm with and without label propagation on standard SemEval all-words tasks using WordNet as the inventory.  
Our proposed algorithms achieve state-of-art performance on many SemEval all-words WSD tasks. In order to assess the effects of training corpus size and language model capacity we also evaluate our algorithms using the New Oxford American Dictionary (NOAD) inventory with SemCor~\cite{miller1993semantic} or MASC~\footnote{http://www.anc.org/MASC/About.html/}. 

\subsection{SemEval Tasks}
In this section, we study the performance of our classifiers on Senseval2~\cite{senseval2}, Senseval3~\cite{senseval3}, SemEval-2007~\cite{senseval7}, SemEval-2013 Task 12~\cite{semeval2013} and SemEval-2015 task 13~\cite{semeval2015}~\footnote{We mapped all senses to WordNet3.0 by using maps in https://wordnet.princeton.edu/wordnet/download/current-version/ and http://web.eecs.umich.edu/~mihalcea/downloads.html}. We focus the study on all-words WSD tasks. 
For a fair comparison with related works, the classifiers are evaluated on all words (both polysemous and monosemous). 

Following related works, we use SemCor or OMSTI~\cite{taghipour2015one} for training. 
In our LP classifiers, unlabeled data for each lemma consists either of $1000$ sentences which contain the lemma, randomly sampled from the web, or all OMSTI sentences (without labels) which contain the lemma.

\begin{table*}[ht]
\centering
\small
\begin{tabular}{llp{0.6cm}p{0.6cm}p{0.6cm}p{0.6cm}p{0.6cm} p{0.6cm}p{0.6cm}p{0.6cm}p{0.6cm}p{0.6cm}p{0.6cm}}
\toprule
 {}&\multicolumn{2}{c}{Senseval2}&\multicolumn{2}{c}{Senseval3}&\multicolumn{2}{c}{SemEval7}&\multicolumn{2}{c}{SemEval7-Coarse}&\multicolumn{1}{c}{SemEval13}\\
\cmidrule(lr){2-3} \cmidrule(lr){4-5} \cmidrule(lr){6-7} \cmidrule(lr){8-9}
 {\bf model} &{\bf all}& {\bf n.} &{\bf all}& {\bf n.} &{\bf all}& {\bf n.} &{\bf all}& {\bf n.} & {\bf n.}  \\\hline
IMS + Word2Vec (T:SemCor)      & 0.634 & 0.742 & 0.653 & 0.701 & 0.578 & 0.686 &       &       &       \\
IMS + Word2Vec (T:OMSTI)       & 0.683 & 0.777 & 0.682 & 0.741 & 0.591 & 0.715 &       &       &       \\
Taghipour and Ng~\shortcite{wsd-word-embedding}      &       &       & 0.682 &       &       &       &       &       &       \\
Chen et al.~\shortcite{chen2014unified}                  &       &       &       &       &       &       & 0.826 & {\bf0.853} &       \\
Weissenborn et al.~\shortcite{weissenborn2015multi}             &       &       &       & 0.688 &       & 0.660 &       & {\bf0.855} & {\bf0.728} \\\hline\hline
Word2Vec (T:SemCor)          & 0.678 & 0.737 & 0.621 & 0.714 & 0.585 & 0.673 & 0.795 & 0.814 & 0.661 \\
LSTM (T:SemCor)              & 0.736 & 0.786 & 0.692 & 0.723 & 0.642 & {\bf0.723} & 0.828 & 0.834 & 0.670 \\
LSTM (T:OMSTI)               & 0.724 & 0.777 & 0.643 & 0.680 & 0.607 & 0.673 & 0.811 & 0.820 & 0.673 \\
LSTMLP (T:SemCor, U:OMSTI)   & 0.739 & 0.797 & 0.711 & 0.748 & {\bf0.637} & 0.704 & {\bf0.843} & 0.834 & 0.679 \\
LSTMLP (T:SemCor, U:1K)      & 0.738 & 0.796 & {\bf0.718} & {\bf0.763} & 0.635 & 0.717 & 0.836 & 0.831 & 0.695 \\
LSTMLP (T:OMSTI, U:1K)       & {\bf0.744} & {\bf0.799} & 0.710 & 0.753 & 0.633 & 0.717 & 0.833 & 0.825 & 0.681 \\\hline
\end{tabular}
\caption{F1 scores on SemEval all-words tasks. T:SemCor stands for models trained with SemCor. U:OMSTI stands for using OMSTI as unlabeled sentences in semi-supervised WSD. IMS + Word2Vec scores are from~\cite{supervised_word2vec}}\label{tab:sem_eval}
\end{table*}

Table~\ref{tab:sem_eval} shows the Sem-Eval results. 
Our proposed algorithms achieve the highest all-words F1 scores except for Sem-Eval 2013. 
Weissenborn et al.\shortcite{weissenborn2015multi} only disambiguates nouns, and it outperforms our algorithms on Sem-Eval 2013 by $4$\%, but is ranked behind our algorithms on Senseval-3 and SemEval-7 tasks with an F1 score more than $6$\% lower than our algorithms. 
Unified WSD~\cite{chen2014unified} has the highest F1 score on Nouns (Sem-Eval-7 Coarse), but our algorithms outperform Unified WSD on other part-of-speech tags. 

\paragraph{Settings}For a fair comparison of Word2Vec and LSTM, we do not use pre-trained word-embeddings as in~\cite{supervised_word2vec}, but instead  train the Word2Vec and LSTM models on a $100$ billion word news corpus~\footnote{The training corpus could not be released, but we have plans to open source the well-trained models} with a vocabulary of the most frequent 1,000,000 words. Our self-trained word embeddings have similar performance to the pre-trained embeddings, as shown in Table~\ref{tab:sem_eval}.
The Word2Vec word vectors are of dimension ${1024}$. The LSTM model has $2048$ hidden units, and inputs are ${512}$-dimensional word vectors. We train the LSTM model by minimizing sampled softmax loss~\cite{sampled_softmax} with Adagrad~\cite{adagrad}. The learning rate is $0.1$. We experimented with other learning rates, and observed no significant performance difference after the training converges. We also downsample frequent terms in the same way as~\cite{mikolov2013efficient}.

\paragraph{Word2Vec vectors Vs. LSTM}To better compare LSTM with word vectors we also build a nearest neighbor classifier using Word2Vec word embeddings and SemCor example sentences, Word2Vec (T:SemCor). It performs similar to  IMS + Word2Vec (T:SemCor), a SVM-based classifier studied in~\cite{supervised_word2vec}.  Table~\ref{tab:sem_eval} shows that the LSTM classifier outperforms the Word2Vec classifier across the board.

\paragraph{SemCor Vs. OMSTI}Contrary to the results observed in~\cite{supervised_word2vec}, the LSTM classifier trained with OMSTI performs worse than that trained with SemCor. It seems that the larger size of the OMSTI training data set is more than offset by noise present in its automatically generated labels. While the SVM classifier studied in~\cite{supervised_word2vec} may be able to learn a model which copes with this noise, our naive nearest neighbor classifiers do not have a learned model and deal less well with noisy labels.

\paragraph{Label propagation}We use the implementation of DIST\_EXPANDER~\cite{ravi2016}. We test the label propagation algorithm with SemCor or OMSTI as labeled data sets and OMSTI or $1000$ random sentences from the web per lemma as unlabeled data. The algorithm performs similarly on the different data sets.

Table~\ref{tab:sem_eval_2015} shows the results of Sem-Eval 2015. The LSTM LP classifier with an LSTM language model achieves the highest scores on nouns and adverbs as well as overall F1. 
The LSTM classifier has the highest F1 on verbs.

\begin{table}[ht]
\centering
\small
\begin{tabular}{p{4cm}lllll }
\toprule
{\bf algorithm}& {\bf all} &{\bf n.} &{\bf v.} &{\bf adj.} &{\bf adv.}  \\\hline
LIMSI & 0.647 & {} & {} & {} & 0.795 \\
DFKI & {} & 0.703 & 0.577 & {} & {} \\ 
UNIBA & {} & {} & {} & 0.790 & {} \\ 
BFS Baseline  & 0.663 & 0.667 & 0.551 & {\bf 0.821} & 0.825 \\\hline\hline
Word2Vec (T:SemCor)          & 0.667 & 0.661 & 0.555 & 0.789 & 0.810 \\
LSTM (T:SemCor)              & 0.721 & 0.713 & {\bf0.642} & 0.813 & 0.845 \\
LSTMLP (T:SemCor, U:1K)      & {\bf0.726} & {\bf0.728} & 0.622 & 0.813 & {\bf0.857} \\ \hline
\end{tabular}
\caption{F1 Scores of SemEval-2015 English Dataset. The BFS baseline uses BabelNet first sense. }\label{tab:sem_eval_2015}
\end{table}

\subsection{NOAD Eval}
\label{sec:noad-evel}
Many dictionary lemmas and senses have no examples in SemCor or OSTMI, giving rise to losses in all-words WSD when these corpora are used as training data. The above SemEval scores do not distinguish errors caused by missing training data for certain labels or inaccurate classifier. 
To better study the proposed algorithms, we train the classifiers with the New Oxford American Dictionary (NOAD)~\cite{NOAD}, in which there are example sentences for each word sense.

\subsubsection{Word Sense Inventory}
\label{sec:inventory}
The NOAD focuses on American English and is based on the Oxford Dictionary of English (ODE)~\cite{ODE}.  It distinguishes between coarse ({\em core}) and fine-grained ({\em sub}) word senses in the same manner as ODE.  Previous investigations~\cite{Navigli:2006:MCS,semeval2007-coarse} using the ODE have shown that coarse-grained word senses induced by the ODE inventory address problems with WordNet's fine-grained inventory, and that the inventory is useful for word sense disambiguation.  

For our experiments, we use NOAD's core senses, and we also use lexicographer-curated example sentences from the Semantic English Language Database (SELD)\footnote{http://oxfordgls.com/our-content/english-language-content/}, provided by Oxford University Press.  We manually annotated all words of the English language SemCor corpus and MASC corpora with NOAD word senses in order to evaluate performance~\footnote{https://research.google.com/research-outreach.html\#/research-outreach/research-datasets}. 
Table~\ref{tab:NOAD} shows the total number of polysemes (more than one core sense) and average number of senses per polyseme in NOAD/SELD (hereafter, NOAD), SemCor and MASC. 
Both SemCor and MASC individually cover around $45\%$ of NOAD polysemes and $62\%$ of senses of those polysemes.
\begin{table}[ht]
\small
\centering
\begin{tabular}{rlcccc}
\toprule
{} & { }  & noun & verb & adj. &adv. \\\hline
\multirow{3}{4cm}{Number of polysemous lemmas in dictionary / corpus}
& NOAD		&  8097	&  2124 	&  2126 	&  266 \\
& SemCor 	&  2833	&  1362 	&  911 	&  193 \\
& MASC 		&  2738	&  1250 	&  829 	&  181 \\\hline\hline
\multirow{3}{4cm}{Sense count per polyseme}
& NOAD 		& 2.46	&  2.58 	&  2.30 	&  2.47 \\
& SemCor 	& 1.44	&  1.69 	&  1.49 	&  1.84 \\
& MASC 		& 1.48	&  1.66 	&  1.48 	&  2.01 \\\hline\hline
\end{tabular}
\caption{NOAD polysemous lemma in NOAD, SemCor and MASC}\label{tab:NOAD}
\end{table}

Table~\ref{tab:data} gives the number of labeled sentences of these datasets. 
Note that although NOAD has more labeled sentences than SemCor or MASC, the average numbers of sentences per sense of these datasets are similar.  This is because NOAD has labeled sentences for each word sense, whereas SemCor (MASC) only covers a subset of lemmas and senses (Table~\ref{tab:NOAD}). 
The last column of Table~\ref{tab:data} shows that each annotated word in SemCor and MASC has an average of more than $4$ NOAD corse senses. Hence, a random guess will have a precision around $1/4$.

\begin{table}[ht]
\small
\centering
\begin{tabular}{cccccc c c}
\toprule
{}&\multicolumn{5}{c}{example count (in 1000's)}&\multirow{2}{2cm}{example count per sense}& \multirow{2}{3cm}{number of candidate senses per example}\\
{dataset} & all & n. & v. & adj. &adv 	& {} & {} \\\hline
NOAD &  {580} &  {312} &  {150} &  {95} &  {13}  &18.43  &  {3.1} \\
SemCor & {115} &  {38} &  {57} &  {11.6} &  {8.6} &14.27  &  {4.1} \\
MASC & {133} &  {50} &  {57} &  {12.7} &  {13.6} &17.38  &  {4.2} \\\hline
\end{tabular}
\caption{Number of examples in each dataset and the average sense count per example.}\label{tab:data}
\end{table}

In the default setting, we use NOAD example sentences as labeled training data and evaluate on SemCor and MASC. We evaluate all polysemous words in the evaluation corpus.

\subsubsection{LSTM classifier}
\label{sec:exp-noad-lstm}
We compare our algorithms with two baseline algorithms:
\begin{itemize}[noitemsep,nolistsep] 
\item Most frequent sense: Compute the sense frequency (from a labeled corpus) and label word $w$ with $w$'s most frequent sense.
\item Word2Vec: a nearest-neighbor classifier with Word2Vec word embedding, which has similar performance to cutting-edge algorithms studied in~\cite{supervised_word2vec} on SemEval tasks. 
\end{itemize}

\begin{table*}[ht]
\centering
\small
\begin{tabular}{llp{0.6cm}p{0.6cm}p{0.6cm}p{0.6cm}p{1cm} p{0.6cm}p{0.6cm}p{0.6cm}p{0.6cm}p{1cm}}
\toprule
 {\bf eval data}&\multicolumn{5}{c}{SemCor}&\multicolumn{5}{c}{MASC}\\
\cmidrule(lr){1-2} \cmidrule(lr){3-7} \cmidrule(lr){8-12} 
 {\bf model} &{\bf train data}& {\bf all} &{\bf n.} &{\bf v.} &{\bf adj.} &{\bf adv.} & {\bf all} &{\bf n.} &{\bf v.} &{\bf adj.} &{\bf adv.} \\\hline
Frequent Sense & SemCor &  	& 	& 	& 	& 	&  0.753	&  0.799	&  0.713	&  0.758	&  0.741 \\
Frequent Sense & MASC &   0.752	&  0.751	&  0.749	&  0.737	&  0.789	& 	& 	& 	& 	&  \\
Word2Vec & NOAD &   0.709	&  0.783	&  0.657	&  0.736	&  0.693	&  0.671	&  0.790	&  0.562	&  0.724	&  0.638 \\
Word2Vec & SemCor &  	& 	& 	& 	& 	&  0.692	&  0.806	&  0.592	&  0.754	&  0.635 \\
Word2Vec & NOAD,SemCor &  	& 	& 	& 	& 	&  0.678	&  0.808	&  0.565	&  0.753	&  0.604 \\
Word2Vec & MASC &   0.698	&  0.785	&  0.619	&  0.766	&  0.744	& 	& 	& 	& 	&  \\
Word2Vec & NOAD,MASC &   0.695	&  0.801	&  0.605	&  0.767	&  0.719	& 	& 	& 	& 	&  \\
\cmidrule(lr){2-12}
LSTM & NOAD &   0.786	&  0.796	&  0.782	&  0.781	&  0.784	&  0.786	&  0.805	&  0.772	&  0.776	&  0.786 \\
LSTM & SemCor &  	& 	& 	& 	& 	&  0.799	&  0.843	&  0.767	&  0.808	&  0.767 \\
LSTM & NOAD,SemCor &  	& 	& 	& 	& 	&  {\bf0.812}	& {\bf0.846}	&  {\bf0.786}	&  {\bf0.816}	&  {\bf0.798} \\
LSTM & MASC &   0.810	&  0.825	&  0.799	&  0.809	&  {\bf0.825}	& 	& 	& 	& 	&  \\
LSTM & NOAD,MASC &   {\bf0.821}	&  {\bf0.834}	&  0.814	&  {\bf0.818}	&  0.821	& 	& 	& 	& 	&  \\
\hline \hline
\end{tabular}
\caption{F1 scores of LSTM algorithm in comparison with baselines}\label{tab:baseline}
\end{table*}

\begin{table*}[ht]
\centering
\small
\begin{tabular}{llp{0.6cm}p{0.6cm}p{0.6cm}p{0.6cm}p{1cm} p{0.6cm}p{0.6cm}p{0.6cm}p{0.6cm}p{1cm}}
\toprule
 {\bf eval data}&\multicolumn{5}{c}{SemCor}&\multicolumn{5}{c}{MASC}\\
\cmidrule(lr){1-2} \cmidrule(lr){3-7} \cmidrule(lr){8-12} 
 {\bf model} &{\bf train data}& {\bf all} &{\bf n.} &{\bf v.} &{\bf adj.} &{\bf adv.} & {\bf all} &{\bf n.} &{\bf v.} &{\bf adj.} &{\bf adv.} \\
\cmidrule(lr){2-12}
LSTM & NOAD &   0.769	&  0.791	&  0.759	&  0.751	&  0.672	&  0.780	&  0.791	&  0.768	&  0.780	&  0.726 \\
LSTM & SemCor &  	& 	& 	& 	& 	&  0.656	&  0.663	&  0.668	&  0.643	&  0.581 \\
LSTM & NOAD,SemCor &  	& 	& 	& 	& 	&\bf0.796	&0.805	&\bf0.790	&\bf0.794	&\bf0.742 \\
LSTM & MASC &   0.631	&  0.653	&  0.606	&  0.617	&  0.600	& 	& 	& 	& 	&  \\
LSTM & NOAD,MASC & \bf0.782	&\bf0.803	&\bf0.774	&\bf0.761	&\bf0.688	& 	& 	& 	& 	&  \\
\hline \hline
\end{tabular}
\caption{Macro F1 scores of LSTM classifier}\label{tab:baseline-macro}
\end{table*}

Table~\ref{tab:baseline} compares the F1 scores of the LSTM and baseline algorithms. 
LSTM outperforms Word2Vec by more than $10\%$ over all words, where most of the gains are from verbs and adverbs. The results suggest that syntactic information, which is well modeled by LSTM but ignored by Word2Vec, is important to distinguishing word senses of verbs and adverbs. 

\paragraph{Change of training data}
By default, the WSD classifier uses the NOAD example sentences as training data. We build a larger training dataset by adding labeled sentences from SemCor and MASC, and study the change of F1 scores in Table~\ref{tab:baseline}.  
Across all part of speech tags and datasets, F1 scores increase after adding more training data.
We further test our algorithm by using SemCor (or MASC) as training data (without NOAD examples). The SemCor (or MASC) trained classifier is on a par with the NOAD trained classifier on F1 score. However, the macro F1 score of the former is much lower than the latter, as shown in Table~\ref{tab:baseline-macro}, because of the limited coverage of rare senses and words in SemCor and MASC.

\paragraph{Change of language model capacity}

\begin{figure*}[ht]
    \centering
    \begin{subfigure}[b]{0.48\textwidth}
        \centering
        \includegraphics[width=0.9\textwidth]{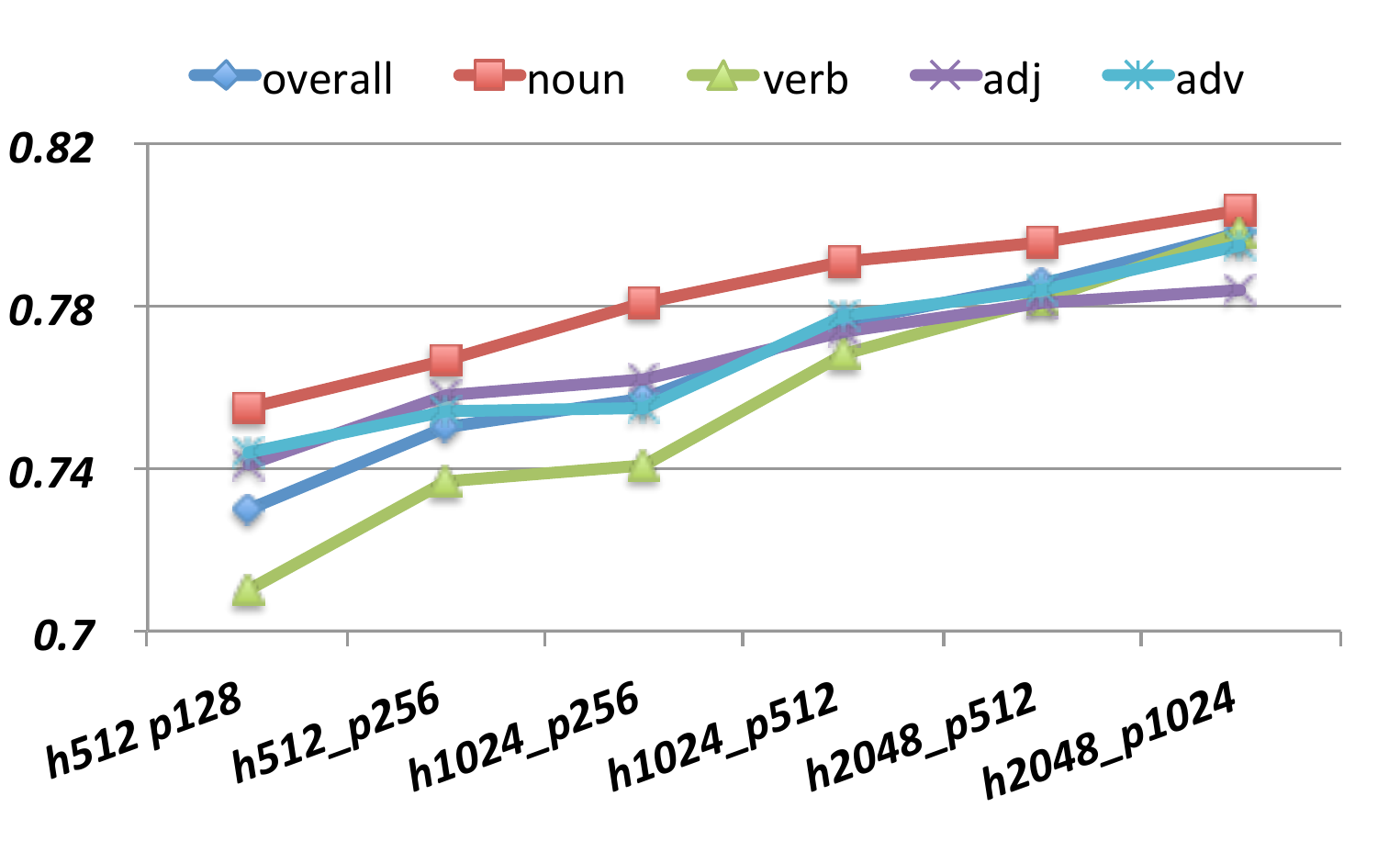}
        \caption{eval on SemCor}
        \label{fig:dim_semcor}
    \end{subfigure}
    \hfill
    \begin{subfigure}[b]{0.48\textwidth}
        \centering
        \includegraphics[width=0.9\textwidth]{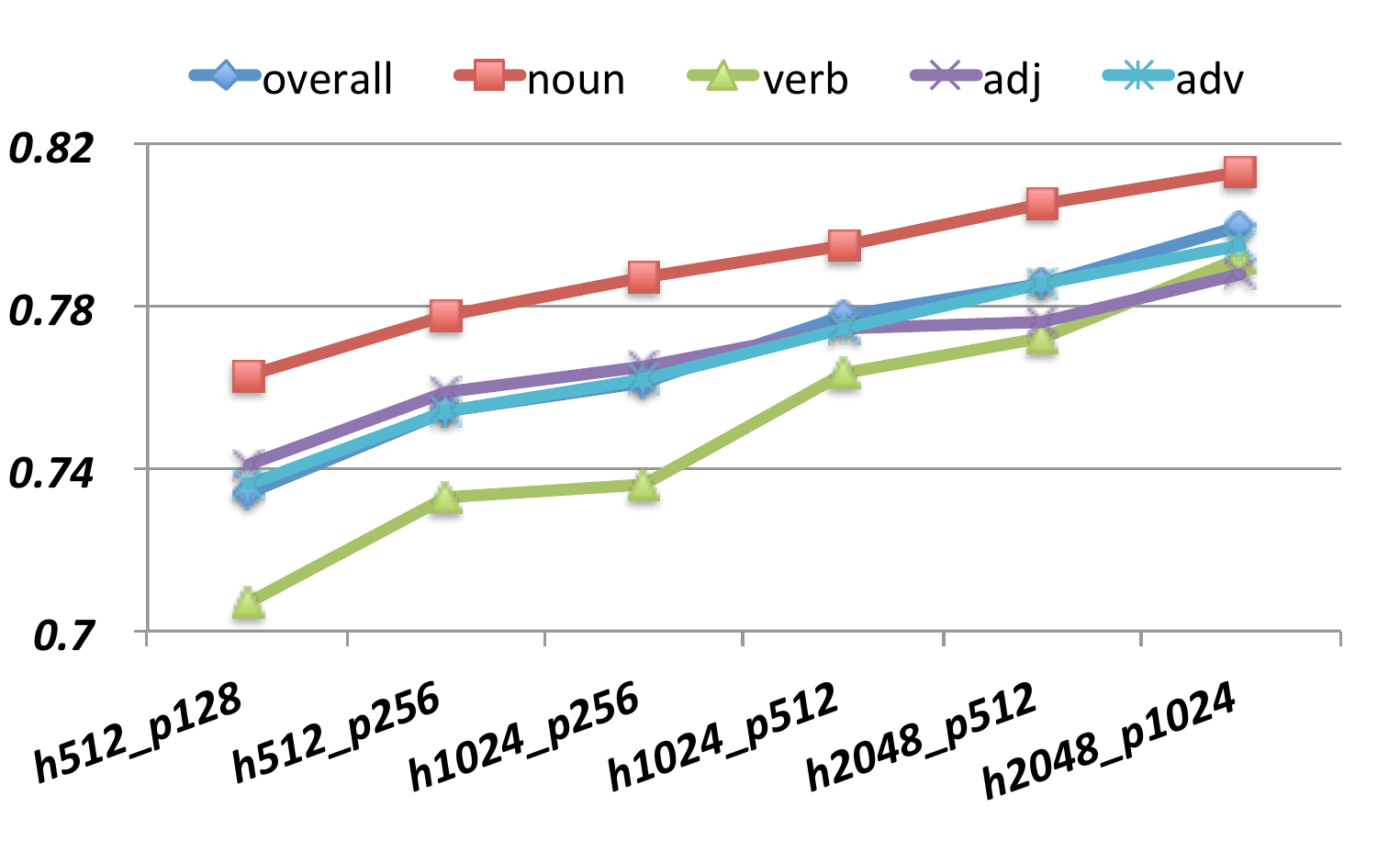}
        \caption{eval on MASC}
        \label{fig:dim_masc}
    \end{subfigure}
    \hfill
    \caption{F1 scores of LSTM models with different capacity: h is the number of hidden units; p is the embedding dimension.}
    \label{fig:dim}
\end{figure*}

In this experiment, we change the LSTM model capacity by varying the number of hidden units $h$ and the dimensions of the input embeddings $p$ and measuring F1. 
Figure~\ref{fig:dim} shows strong positive correlation between F1 and the capacity of the language model. However, larger models are slower to train and use more memory. 
To balance the accuracy and resource usage, we use the second best LSTM model ($h = 2048$ and $p = 512$) by default.

\subsubsection{Semi-supervised WSD}
We evaluate our semi-supervised WSD classifier in this subsection. 
We construct the graph as described in Section~\ref{sec:semi-supervised-wsd} and run LP to propagate sense labels from the seed vertices to the unlabeled vertices. We
evaluate the performance of the algorithm by comparing the predicted labels and the gold labels on eval nodes. 
As can be observed from Table~\ref{tab:semi-supervised},  LP did not yield clear benefits when using the Word2Vec language model.
We did see significant improvements, 6.3\% increase on SemCor and 7.3\% increase on MASC, using LP with the LSTM language model. 
We hypothesize that this is because LP is sensitive to the quality of the graph distance metric.

\begin{table*}[ht]
\centering
\small
\begin{tabular}{llp{0.6cm}p{0.6cm}p{0.6cm}p{0.6cm}p{1cm} p{0.6cm}p{0.6cm}p{0.6cm}p{0.6cm}p{1cm}}
\toprule
 {\bf eval data}&\multicolumn{5}{c}{SemCor}&\multicolumn{5}{c}{MASC}\\
\cmidrule(lr){1-2} \cmidrule(lr){3-7} \cmidrule(lr){8-12} 
 {\bf model} &{\bf train data}& {\bf all} &{\bf n.} &{\bf v.} &{\bf adj.} &{\bf adv.} & {\bf all} &{\bf n.} &{\bf v.} &{\bf adj.} &{\bf adv.} \\
\cmidrule(lr){2-12}
Word2Vec LP & NOAD &   0.642	&  0.733	&  0.554	&  0.705	&  0.725	&  0.643	&  0.752	&  0.524	&  0.726	&  0.664 \\
\cmidrule(lr){2-12}
LSTM LP & NOAD &   0.822	&  0.859	&  0.800	&  0.817	&  0.816	&  0.831	&  0.865	&  0.806	&  0.825	&  0.821 \\
LSTM LP & NOAD,SemCor &  	& 	& 	& 	& 	&\bf0.872	&\bf0.897	&\bf0.852	&\bf0.865	&\bf0.868 \\
LSTM LP & NOAD,MASC & \bf0.873	&\bf0.883	&\bf0.870	&\bf0.858	&\bf0.874	& 	& 	& 	& 	&  \\
\hline \hline
\end{tabular}
\caption{F1 scores of label propagation}\label{tab:semi-supervised}
\end{table*}

\paragraph{Change of seed data: }
As can be seen in Table~\ref{tab:semi-supervised}, LP substantially improves classifier F1 when the training datasets are SemCor+NOAD or MASC+NOAD. 
As discussed in Section~\ref{sec:semi-supervised-wsd}, the improvement may come from explicitly modeling the sense prior.
We did not see much performance lift by increasing the number of unlabeled sentences per lemma. 

\paragraph{Change of graph density: }
By default, we construct the LP graph by connecting two nodes if their affinity is above $95$\% percentile. 
We also force each node to connect to at least $10$ neighbors to prevent isolated nodes. 
Table~\ref{tab:density} shows the performance of the LP algorithm by changing the percentile threshold. 
The F1 scores are relatively stable when the percentile ranges between $85$ to $98$, but decrease when the percentile drops to $80$. 
Also, it takes longer to run the LP algorithm on a denser graph. We pick the $95$ percentile in our default setting to achieve both high F1 scores and short running time. 

\begin{table*}[ht]
\centering
\small
\begin{tabular}{lp{0.6cm}p{0.6cm}p{0.6cm}p{0.6cm}p{0.6cm}p{0.6cm} p{0.6cm}p{0.6cm}p{0.6cm}p{0.6cm}p{0.6cm}p{0.6cm}}
\toprule
 {}&\multicolumn{6}{c}{SemCor}&\multicolumn{6}{c}{MASC}\\
 \cmidrule(lr){2-7} \cmidrule(lr){8-13} 
 {\bf pos-tag} & {\bf 98} &{\bf 95} &{\bf 90} &{\bf 85} &{\bf 80} & {\bf 70} &{\bf 98} &{\bf 95} &{\bf 90} &{\bf 85} &{\bf 80} & {\bf 70}  \\
\hline \hline
overall &  {\bf0.823} & 0.822 & {\bf0.823} & 0.818 & 0.813 & 0.800 & 0.827 & 0.831 & {\bf0.835} & 0.830 & 0.824 & 0.806 \\
n. &  0.848 & {\bf0.859} & 0.852 & 0.846 & 0.840 & 0.828 & 0.863 & 0.865 & {\bf0.868} & 0.865 & 0.861 & 0.847 \\
v. &  {\bf0.810} & 0.800 & 0.803 & 0.797 & 0.792 & 0.778 & 0.800 & {\bf0.806} & {\bf0.806} & 0.799 & 0.794 & 0.769 \\
\hline
\end{tabular}
\caption{F1 scores of the LSTM LP trained on NOAD with varying graph density.}\label{tab:density}
\end{table*}

\section{Conclusions and Future Work}
\label{sec:discussion}
In this paper, we have presented two WSD algorithms which combine (1) LSTM neural network language models trained on a large unlabeled text corpus,  with (2) labeled data in the form of example sentences, and, optionally, (3) unlabeled data in the form of additional sentences. Using an LSTM language model gave better performance than one based on Word2Vec embeddings. The best performance was achieved by our semi-supervised WSD algorithm which builds a graph containing labeled example sentences augmented with a large number of unlabeled sentences from the web, and classifies by propagating sense labels through this graph.

Several unanswered questions suggest lines of future work. 
Since our general approach is amenable to incorporating any language model, further developments in NNLMs may permit increased performance.  We would also like to better understand the {\em limitations} of language modeling for this task: we expect there are situations -- e.g., in idiomatic phrases -- where per-word predictions carry little information.

We believe our model should generalize to languages other than English, but have not yet explored this.  Character-level LSTMs~\cite{kim2015character} may provide robustness to morphology and diacritics and may prove useful even in English for spelling errors and out of vocabulary words.

We would like to see whether our results can be improved by incorporating global (document) context and multiple embeddings for polysemous words~\cite{huang2012improving}.

Finally, many applications of WSD systems for nominal resolution require integration with resolution systems for named entities, since surface forms often overlap~\cite{moro2014entity,babelnet}.  This will require inventory alignment work and model reformulation, since we currently use no document-level, topical, or knowledge-base coherence features.

We thank our colleagues and the anonymous reviewers for their insightful comments on this paper.

\bibliography{coling2016}
\bibliographystyle{acl}

\end{document}